# An automated method of identifying incorrectly labelled images based on the sequences of loss functions of deep learning networks

Zhipeng Zhang[1], Wenhui Shou[1], Wengting Ma[1], Dongjia Xing[1], Qingqing Xu[1], Li-Qun Xu[1], Qingxia Fan [2], Ling Xu[2]

[1] China Mobile Research Institute, Beijing 100032, China
[2] Shenyang He Eye Hospital, Shenyang 110034, China
shouwenhui@chinamobile.com

**Abstract.** Deep learning has been widely applied to medical image analysis tasks. Since the labelled medical images are the foundation of the training, validation, and test of deep learning classification models, the quality of labelling process could directly affect the performance of the models. However, it was estimated that up to ten percent of manually labelled medical images may be incorrectly labelled. In this paper, by utilizing the sequences of loss functions of deep learning classification networks through multiple training epochs, an automated method of identifying incorrectly labelled medical images was proposed. For those identified images, their labels could be further reviewed and updated by senior and experienced physicians, ultimately improving the quality of labelled medical image datasets, as well as the performance of the deep learning models.

Two experiments were carried out to validate the effectiveness of the proposed method, based on a specific fundus image dataset for referable diabetic retinopathy screening. a) In the first experiment, the effectiveness of the method to accurately identify the incorrectly labelled samples from the whole labelled dataset was verified. For a fundus image dataset comprising 10788 samples with gold-standard labels (5394 non- referable diabetic retinopathy samples and 5384 referable diabetic retinopathy samples), the labels of a small part (6%, 648) of the images were intentionally changed to the opposite, in order to simulate the real-world situation. By utilizing the proposed method, 75.31% (488) of the incorrectly labelled samples were successfully identified, and only 4.85% (492) of the correctly labelled samples were wrongly identified as the incorrectly labelled ones. b) In the second experiment, by further reviewing those 980 samples (only 9.1% of the whole dataset) that were identified as incorrectly labelled from the dataset and updating their labels to the correct ones, the deep learning classification model for referable diabetic retinopathy screening was retrained. Tested on an independent test dataset with completely correct labels (700 non- referable diabetic retinopathy samples and 700 referable diabetic retinopathy samples), the best accuracy of the model was increased from 95.93% (trained on the dataset with 6% incorrectly labelled samples) to 96.50% (trained on the revised dataset with 1.5% incorrectly labelled samples), approaching the ideal value 96.57% (trained on the original dataset with 0% incorrectly labelled samples), demonstrating the effectiveness of the proposed method to improve the performance of the deep learning models.

# 1 Introduction

Artificial intelligence (AI) has drawn intense interests around the world. In the healthcare industry, self-diagnostic symptom checkers and clinical decision support systems become a research topic and deep learning techniques-enabled medical image analysis has been applied to almost every medical specialty [1-6]. To realize this vision, it is necessary to obtain a large amount of medical images which were dedicatedly labelled by a group of physicians. These images would be used as either the training samples to construct a deep convolutional neural network (CNN) model, or the validation and test samples to verify the performance of the model.

However, when physicians manually label the medical images, especially a large amount of real-world images with varying levels of image quality, which were taken by different types of medical devices in different clinical and screening settings, it is inevitable that there will be incorrectly labelled samples. The main reasons were as follows:

(1) For the medical image labelling tasks, lesions sometimes occupy a small portion of pixels in the images, such as the microaneurysm (Fig 1.a) and indentation in the arteriovenous crossing (Fig 1.b) in fundus images. The physicians may omit them due to fatigue or limited expertise, making the labels of the images incorrect.

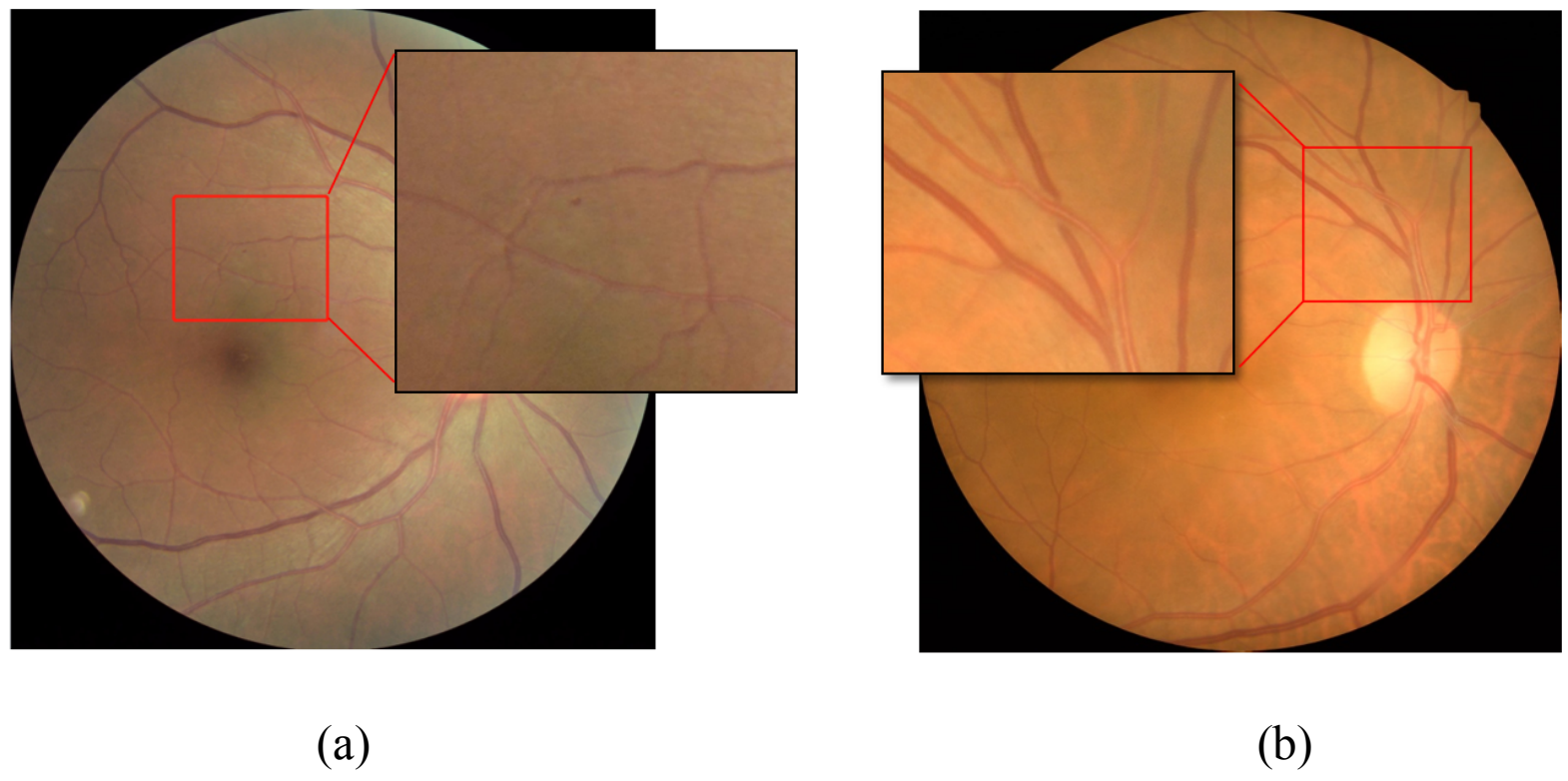

(a) (b)

**Fig. 1.** Fundus images with small lesions such as the microaneurysm and indentation in the arteriovenous crossing

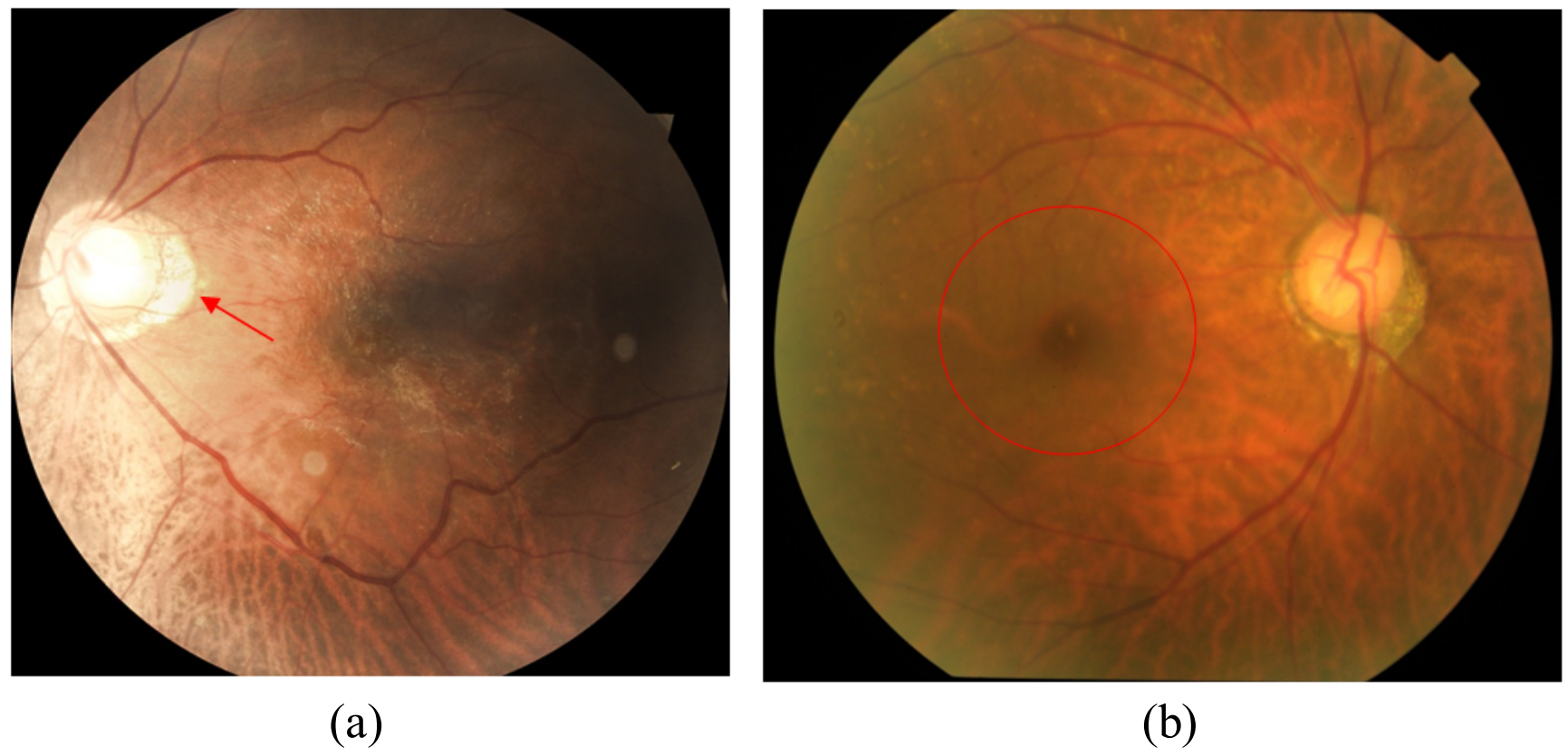

(a) (b)

**Fig. 2.** Fundus images in the intermediate state of high myopia and age-related macular degeneration

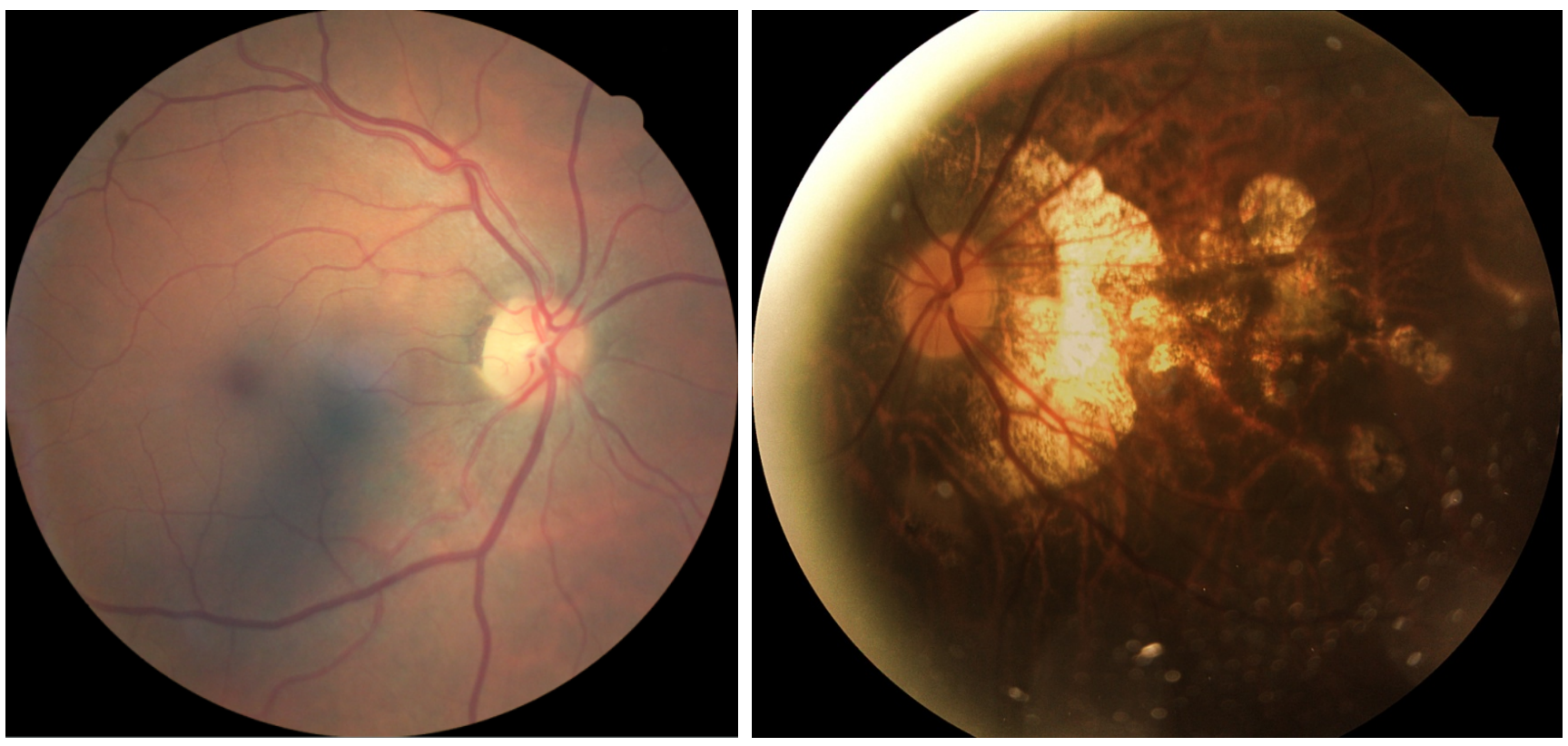

**Fig. 3.** A fundus image hard to be labelled as gradable or un-gradable

(2) The boundaries between the different categories were sometimes illegible in medical images. For example, for the fundus images with high myopia, peripapillary atrophy appears as a middle stage between normal with leopard pattern and visible sclera, which is the clinical feature of pathologic myopia. It is hard for ophthalmologists to agree on whether an image with the peripapillary atrophy (Fig 2.a) should be labelled as pathologic myopia or non-pathologic myopia. Thus, those challenging samples may be marked with opposite labels, reflecting the inconsistent opinions of different ophthalmologists. Another case is the identification of age-related macular degeneration (AMD) in fundus images, where drusen usually starts to appear at 1 optic-disc-diameter away from the center of macular (Fig 2.b). However, it is hard to accurately quantify the distance and make a correct and consistent labelling decision.

(3) In some situations, labelling standards for medical images are not absolutely objective and measurable[7], resulting in poor consistency on those ambiguous images. For example, the standard of labelling a medical image as gradable (good quality) or un-gradable (poor quality) differs from that of other kinds of images, because it is more focused on whether the target lesion is clear and prominently photographed, as shown in Fig 3. So the manual labelling process introduces subjective factors unavoidably.

If the incorrectly labelled images were partitioned into the training set, the model would be misled and the performance would be affected. If they were partitioned into the validation set or the test set, the performance of the model could not be authentically evaluated. In general, it is necessary to look for senior and experienced physicians to review all the initial labelling results to ensure the results' reliability, which is time- and energy-consuming. If the incorrectly labelled samples can be automatically identified by an algorithm, the senior and experienced physicians simply need to review a small portion of the images for corrections, which greatly improves the efficiency of data preparation, lays a reliable foundation for the training, validation and test of the deep learning model, and improve the performance of the model.

In this study, a novel automated method of identifying incorrectly labelled medical images based on the sequences of loss functions of deep learning networks was proposed, and its feasibility and effectiveness were evaluated in a fundus image dataset for diabetic retinopathy classification.

## 2 Methods

In deep learning networks, loss functions measure the discrepancy between the true probabilities and the estimated ones [8]. During the training procedure, optimization algorithms are often used to update the parameters of a model so as to minimize the loss function's value. As the model converges to the optimal position through multiple epochs, the loss value goes down gradually. For those correctly labelled samples, the sequences of loss functions always present a similar decreasing trend. For those incorrectly labelled samples, since their 'ground-truth' were wrong, making the trained model very confused, the sequences of loss functions often show completely different trends from the correctly labelled samples. For example, the loss values of incorrectly labelled samples may maintain at a high level, or present fluctuation changes. Therefore, based on the sequences of loss functions of deep learning networks in multiple training epochs, we can extract the features and cluster all the samples into two categories, one of which represents the correctly labelled samples and the other represents the incorrectly labelled samples.

1) Model training for multiple epochs

The labelled dataset, which contains both correctly and incorrectly labelled samples, was used to train a deep learning network pre-trained on ImageNet dataset[9] through N (N>2) epochs for the target disease classification task. Before the training process, all the images can be resized to a predefined size and data augmentation methods can be used to promote the diversity of the dataset. For each training epoch, we saved each sample's loss value based on the loss function, i.e. cross-entropy loss, and the learning rate was 5e-5. Thus a sequence of loss function with a length of N was obtained for each sample.

2) Feature extraction

As mentioned above, the sequences of correctly labelled samples always present similar decreasing trends, thus six features were extracted to describe the characteristics of the trends.

F1: The mean value of the sequence (with a length of N), representing the global level of the sample's loss;

F2: The mean value of the first P (1<=P<N) terms of the sequence, representing the local level of the sample's loss during the early stage of model training;

F3: The position Q (1<=Q<N) in the sequence, which experiences the biggest decrease level between the Qth term and the (Q+1)th term;

F4: The mean value of the first Q terms of the sequence, representing the local level of the sample's loss before the significant decrease;

F5: The mean value of the last N-Q terms of the sequence, representing the local level of the sample's loss after the significant decrease;

F6: F4 minus F5, representing the extent of the significant decrease.

3) Clustering

Based on the six features of each sample in the dataset, k-means clustering method[10] was used to cluster the samples into two categories (k=2), one of which represents the correctly labelled samples and the other represents the incorrectly labelled samples. The stopping criteria of the method includes the maximum number of iterations (300), and the relative tolerance with regards to inertia to declare convergence (1e-4). Those incorrectly labelled samples identified by the method could be further reviewed by senior and experienced physicians, in order to improve the quality of the labelled dataset and the performance of the model.

# 3 Results

The proposed method was validated on a fundus image dataset (denoted as A) for referable diabetic retinopathy (defined as the presence of moderate and worse diabetic retinopathy) screening. Most images of the dataset were obtained using different types of fundus camera during the eye disease screening programs conducted by the He Eye Hospital Group. The rest images were randomly extracted from the training set of Kaggle's Diabetic Retinopathy Detection challenge[11], in order to improve the diversity of the dataset. All the images were labelled and reviewed deliberately by a group of physicians and each image had a ground-truth label, either non- referable diabetic retinopathy (Label = I) or referable diabetic retinopathy (Label = II), as shown in Table 1. The labels were reviewed deliberately by a group of senior and experienced ophthalmologists for more than 3 times, thus each image was considered as a correctly labelled one.

**Table 1.** The label distribution of the gold-standard dataset (A), the synthetic dataset (A'), and the refined dataset (A'')

| Dataset | Label | Number | Correctly labelled | Incorrectly labelled |
|---|---|---|---|---|
| A | Ⅰ(non- referable diabetic retinopathy) | 5394 | 5394 | 0 |
| | Ⅱ(referable diabetic retinopathy) | 5394 | 5394 | 0 |
| | Total | 10788 | 10788 | 0 |
| A' | Ⅰ(non- referable diabetic retinopathy) | 5394 | 5070 | 324 |
| | Ⅱ(referable diabetic retinopathy) | 5394 | 5070 | 324 |
| | Total | 10788 | 10140 | 648 |
| A'' | Ⅰ(non- referable diabetic retinopathy) | 5394 | 5284 | 110 |
| | Ⅱ(referable diabetic retinopathy) | 5394 | 5344 | 50 |
| | Total | 10788 | 10628 | 160 |

In order to simulate the real-world labelled dataset including incorrectly labelled samples, a synthetic dataset (denoted as A') was created, through intentionally changing the labels of a small part (6%) of the images, making those correctly labelled samples become incorrectly labelled.

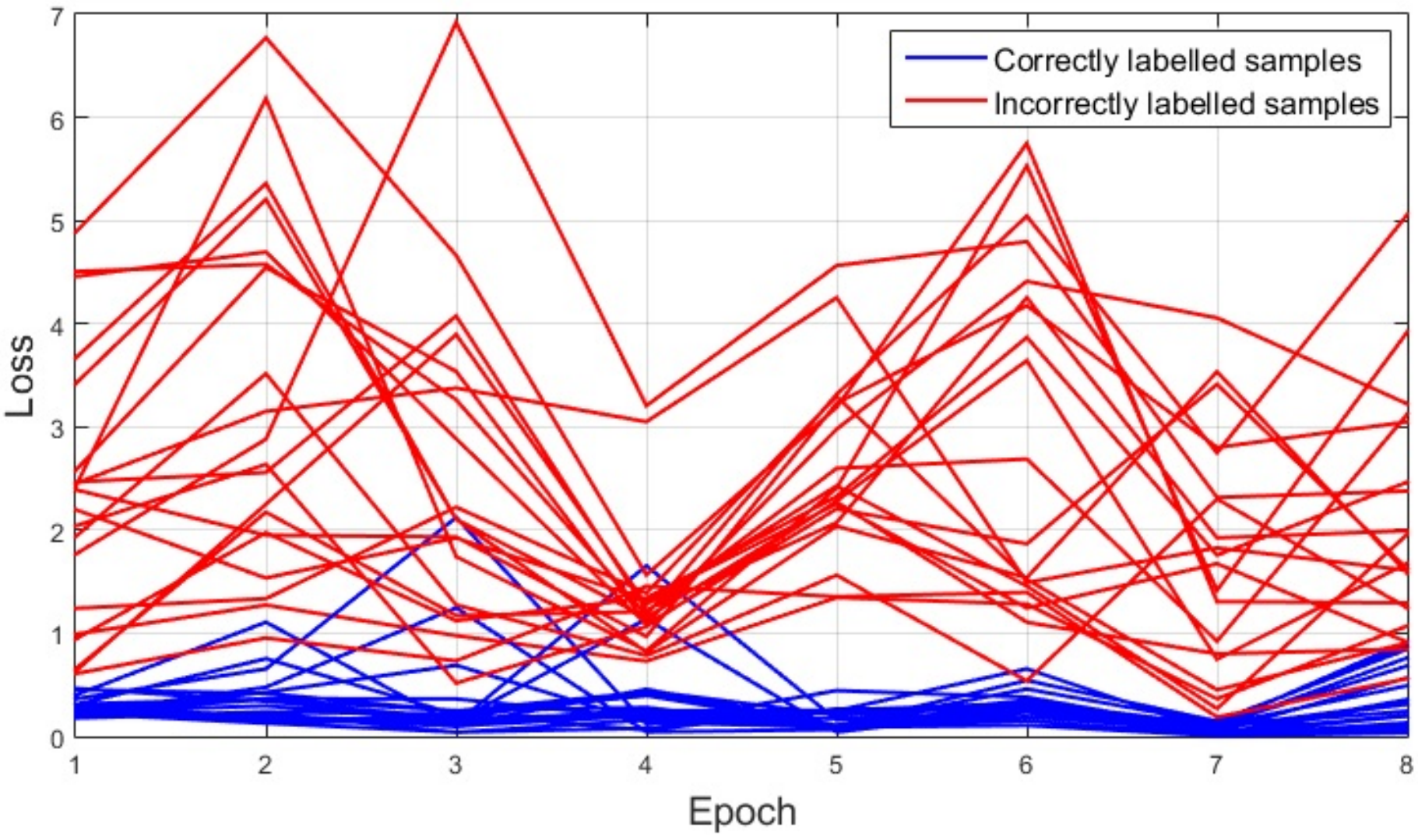

**Fig. 4.** The loss sequences of 20 typical correctly labelled samples in the synthetic dataset A'

In the first experiment, the effectiveness of the method to accurately identify the incorrectly labelled samples from the whole labelled dataset was examined. The synthetic dataset A' was used to train the Inception-v3 network[12] pre-trained on ImageNet dataset and the number of epochs was 8 (N=8). In the pre-processing phase, all images were empirically resized to a predefined 550×550 pixels. Fig 4 shows the sequences of loss function of 20 typical correctly labelled samples and 20 typical incorrectly labelled samples respectively. Six features were further extracted based on the sequence of loss function of each sample (P = 4 for F2) and k-means clustering method was used to cluster the samples into two categories. The sensitivity of incorrectly labelled sample identification was 75.31% (=488/648), and the specificity was 95.15% (=9648/10140).

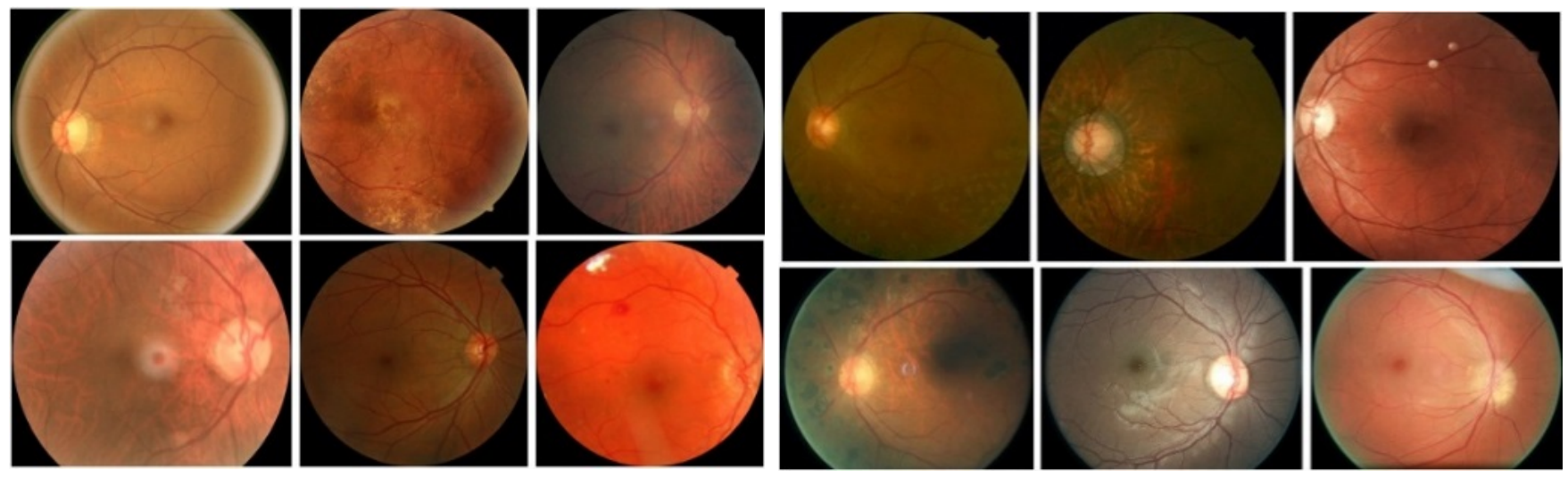

(a) Label II was changed to Label I (b) Label I was changed to Label II

**Fig. 5.** Typical incorrectly labelled samples which were not identified by the method

Typical incorrectly labelled samples which were not identified by the method were shown in Fig 5. Fig 5.a shows six images with referable diabetic retinopathy, which were labelled as non- referable diabetic retinopathy in the dataset A'. However, due to either the poor image quality or the limited area of lesions, they looked similar to those images with non- referable diabetic retinopathy, which means it is difficult to identify them successfully among the samples with non- referable diabetic retinopathy. Fig 5.b shows six images with non- referable diabetic retinopathy, which were labelled as referable diabetic retinopathy in the dataset A'. Because of the noise such as water spots, reflections and lens stain, they looked similar to those images with referable diabetic retinopathy, which means it is difficult to identify them successfully among the samples with referable diabetic retinopathy.

In the second experiment, in order to verify the effectiveness of the proposed method to improve the performance of the deep learning models, the labels of those 980 samples (only 9.1% of the whole dataset) that were identified as incorrectly labelled were reviewed and updated as the correct ones, forming a refined dataset A" with 1.5% incorrectly labelled samples (as shown in Table 1).

Based on the dataset A'', the deep learning classification model for referable diabetic retinopathy screening was retrained using the same network architecture and the performance of the model was compared to the model trained on the dataset A and the model trained on the dataset A'. As shown in Fig 6.a and Fig 6.b, compared to the model trained on A', the accuracy of the model trained on A'' was increased and the average loss was decreased at each training epoch, which was tested on an independ-ent dataset B including 1400 fundus images with completely correct labels (700 images with non- referable diabetic retinopathy and 700 images with referable diabetic retinopathy). During 30 epochs, the best accuracy was increased from 95.93% (model trained on A') to 96.50% (model trained on A''), approaching the ideal value 96.57% (model trained on A), which demonstrates that the automated identification and manually review of the incorrectly labelled samples can improve the performance of the deep learning models.

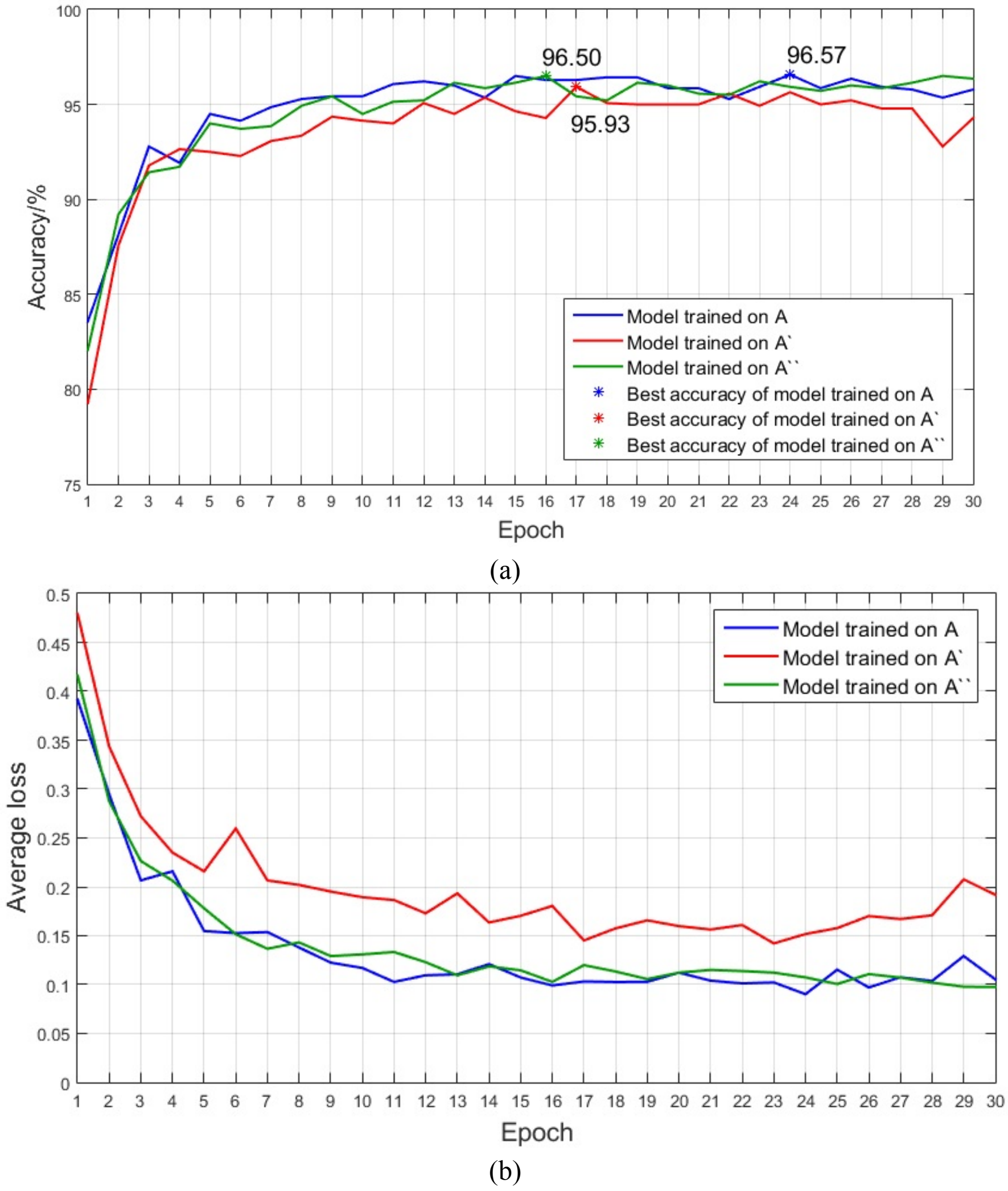

**Fig. 6.** The accuracy and average loss of the models trained on the dataset A, A' and A'' through 30 epochs, tested on an independent dataset B

# 4 Conclusions

An automated method of identifying incorrectly labelled medical images based on the sequences of loss functions of deep learning networks was proposed. Through the accurate identification of those incorrectly labelled samples and the further review procedure, the consistency and reliability of dataset labelling can be effectively improved, which lays a better foundation for network training, and ultimately increases the performance of the deep learning models. Further research is necessary to validate the effectiveness of the method for multi-class classification and segmentation tasks.